# Predictive Modeling of Effluent Temperature in SAT Systems Using Ambient Meteorological Data: Implications for Infiltration Management

Roy Elkayam[†]

[†]Mekorot Water Company, Lincoln Street, Tel-Aviv – Yafo, 6492105, Israel.

The Institute of Chemistry, The Hebrew University of Jerusalem, Jerusalem, 9190401, Israel

**Abstract**

Accurate prediction of effluent temperature in recharge basins is essential for optimizing the Soil Aquifer Treatment (SAT) process, as temperature directly influences water viscosity and infiltration rates. This study develops and evaluates predictive models for effluent temperature in the upper recharge layer of a Shafdan SAT system recharge basin using ambient meteorological data. Multiple linear regression (MLR), neural networks (NN), and random forests (RF) were tested for their predictive accuracy and interpretability. The MLR model, preferred for its operational simplicity and robust performance, achieved high predictive accuracy ($R^2$ = 0.86-0.87) and was used to estimate effluent temperatures over a 10-year period. Results highlight pronounced seasonal temperature cycles and the importance of topsoil temperature in governing the thermal profile of the infiltrating effluent. The study provides practical equations for real-time monitoring and long-term planning of SAT operations.

1. Introduction

Soil aquifer treatment (SAT) systems are widely used for managed aquifer recharge (MAR) in arid and semi-arid regions, offering both water quality improvement and groundwater replenishment (Gharbia et al., 2024; Sharma and Kennedy, 2017). In these systems, treated effluent infiltrates through the unsaturated zone, where it undergoes physical, biological, and chemical transformations before reaching the aquifer (Amy and Drewes, 2007; Bouwer, 2002; Elkayam et al., 2020). SAT systems are particularly valuable in water-scarce regions, with over 60 years of global application demonstrating their effectiveness in sustainable water management (Dillon et al., 2019).

The efficiency of SAT systems is critically dependent on infiltration rates, which are governed by several factors including soil hydraulic properties, effluent quality, and temperature (Bouwer, 2002). Temperature plays a particularly important role through its effect on water viscosity, with studies showing that temperature variations of 10°C can alter infiltration rates by 30-50\% (Braga et al., 2007; Constantz, 1982; Lin et al., 2003). This temperature-viscosity relationship follows the exponential Guzman-Andrade equation (Lin et al., 2003), where viscosity decreases exponentially with increasing temperature, thereby enhancing hydraulic conductivity and infiltration capacity.

Despite the recognized importance of temperature in SAT operations, long-term historical temperature data for the upper layer of recharge basins are often lacking due to limited monitoring infrastructure and the challenging environment of intermittently flooded basins (Heilweil et al., 2006). Traditional temperature monitoring requires expensive, specialized equipment that must withstand repeated wetting-drying cycles and potential damage from basin maintenance activities. This data gap limits the ability of operators to optimize recharge schedules, predict seasonal performance variations, and conduct long-term system planning.

Recent advances in machine learning and data science offer promising solutions for developing predictive models using readily available meteorological data (Ferrario et al., 2025; Markan and Kamboj, 2025). Previous studies have successfully applied various modeling approaches to predict subsurface temperatures in agricultural and hydrological contexts, with ambient air temperature consistently identified as the primary predictor(Biazar et al., 2024; Bilgili, 2010). However, limited research has specifically focused on predicting effluent temperatures in SAT systems, where the thermal dynamics may differ from natural soil systems due to continuous infiltration and the thermal properties of treated wastewater.

This study addresses this research gap by developing and validating predictive models to estimate effluent temperature in a large-scale SAT system's recharge basin using ambient meteorological parameters. The research aims to provide a robust, scalable method for temperature prediction that supports both real-time monitoring and long-term operational planning, ultimately improving the efficiency and management of SAT systems in water reclamation applications.

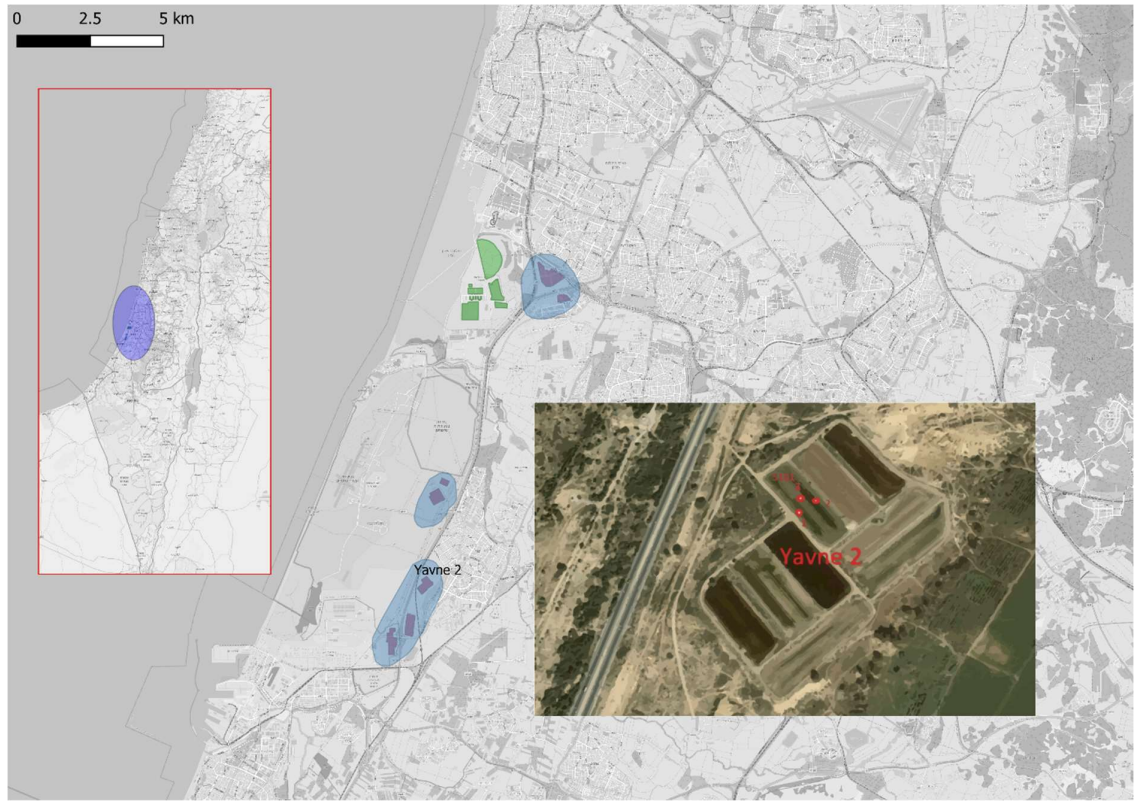

*Figure 1: Map showing the locations of the seven recharge areas at the Shafdan SAT system and the Shafdan Mechanical Biological Plant treatment (MBPT), including an inset map of Israel for geographical context. The map is overlaid with a drone photograph of the Yavne 2 basins, illustrating the location of the 3 Sentek sensors*

2. Methodology

The study focused on Recharge Basin 5101 (Figure 1) within the Shafdan SAT system, Israel's largest wastewater treatment and recharge facility, located approximately 20 km south of Tel Aviv in a Mediterranean climate zone.

## 2.1. Data Collection

**Effluent Temperature Data:** Temperature measurements were collected every 30 minutes over a 21-month period (June 3, 2022, to March 1, 2024) from Recharge Basin 5101. Three Sentek soil moisture and temperature probes (Sentek Technologies, 2025)were installed at distinct locations within the basin as part of a separate study on the unsaturated zone. Each probe recorded temperature at 12 depths (every 10 cm from 5 cm to 115 cm), yielding 36 total measurements per timestep (12 depths X 3 locations).

**Ambient Meteorological Data**: Historical meteorological data, including ambient temperature (T, °C), relative humidity (RH, %), precipitation (P, mm), wind speed (WS, m/s), and wind direction (WD, deg), were retrieved from a nearby weather station operated by the Israeli Meteorological Service (Israel Meteorological Service, 2025). Data were available at 10-minute intervals from January 2015 to March 2025.

## 2.2. Data Preprocessing

**Effluent Temperature:** Drainage Phase Identification: To focus on conditions approximating saturated flow, we selected temperature data only from periods when Recharge Basin 5101 was in the "drainage" phase. This phase begins after flooding, when the basin reaches a target water level and the inlet valve is closed, allowing the effluent to infiltrate into the soil. Drainage periods were identified using operational logs of valve activity and water level records, see also (Elkayam and Lev, 2024).

**Averaging:** (i) Topsoil Temperature: For each 30-minute timestep, we calculated the average temperature at 5 cm depth across the three probe locations. (ii) Profile Average Temperature: We computed the average temperature across all 36 measurements (12 depths × 3 locations) to represent the overall thermal state of the infiltrating effluent.

**Ambient Data:** The 10-minute meteorological data were resampled to 30-minute intervals by averaging to align with the effluent temperature timestep. Outliers in precipitation (P), wind speed (WS), and wind direction (WD) were identified using the interquartile range (IQR) method and excluded due to their excessive variability, leaving T and RH as the primary predictive features.

**Viscosity Calculation:** The temperature dependence of water viscosity was modeled using the Guzman-Andrade equation:

$$(Eq\ 1) \quad \eta(T) = A \cdot e^{\left(\frac{B}{273+T}\right)}$$

where $\eta(T)$ is the dynamic viscosity (Pa·s), $T$ is the temperature (°C), A and B are constants $A = 1.98404 \times 10^{-6}\ Pa \cdot S, B = 1825.85\ C$; values adapted from Lin et al., 2003 (Lin et al., 2003).

### 2.3. Data Analysis

Three machine learning models were evaluated to predict effluent temperature (topsoil and profile average) based on ambient T and RH, implemented using the scikit-learn library (Pedregosa et al., 2011):

**Multiple Linear Regression (MLR)**(Montgomery et al., 2021)**:** A baseline linear model assuming a direct relationship between predictors and effluent temperature.

**Neural Network (NN)**(Goodfellow et al., 2016)**:** A feedforward neural network with two hidden layers (10 and 5 nodes, ReLU activation) to capture non-linear patterns.

**Random Forest (RF)**(Breiman, 2001)**:** An ensemble model with 100 trees, selected as the best-performing model due to its robustness and accuracy.

## 3. Results and Discussion

### 3.1. Model Training and Assessment

Models were trained on 80% of the 21-month dataset, randomly split, and tested on the remaining 20%. Performance was assessed using $R^2$ and root mean square error (RMSE). Feature importance for all models was determined using permutation importance, where the decrease in $R^2$ after randomly shuffling each feature's values quantifies its contribution (Breiman, 2001).

Initially, all four features (i.e., Temperature (T), Relative Humidity (RH), Pressure (P), and Wind Speed (WS)) were utilized to predict effluent temperature. However, the feature importance analysis indicated that the contributions of Pressure and Wind Speed were insignificant. Therefore, only Temperature and Relative Humidity were retained as relevant features.

Figure 2 demonstrates the predictive performance and feature importance of three models: (A) MLR, (B) NN, and (C) RF. The left column displays time series plots comparing actual (blue) and predicted (orange) values for topsoil temperature (Figure 2 I) and profile average temperature (Figure 2 II). Model performance metrics and feature importance values are summarized in Tables 1 and 2 respectively.

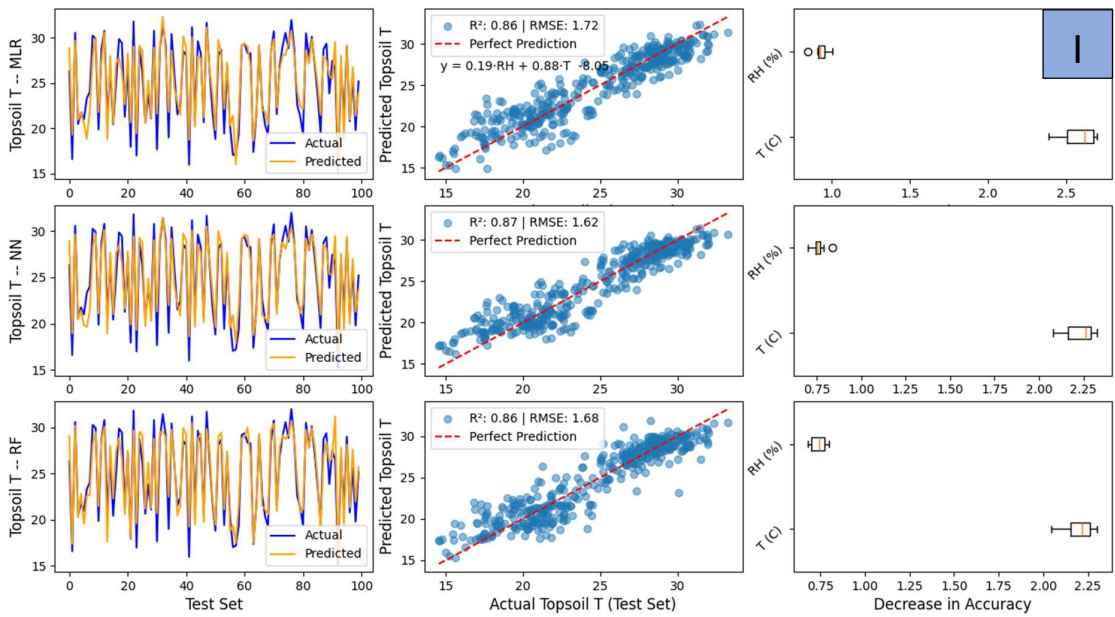

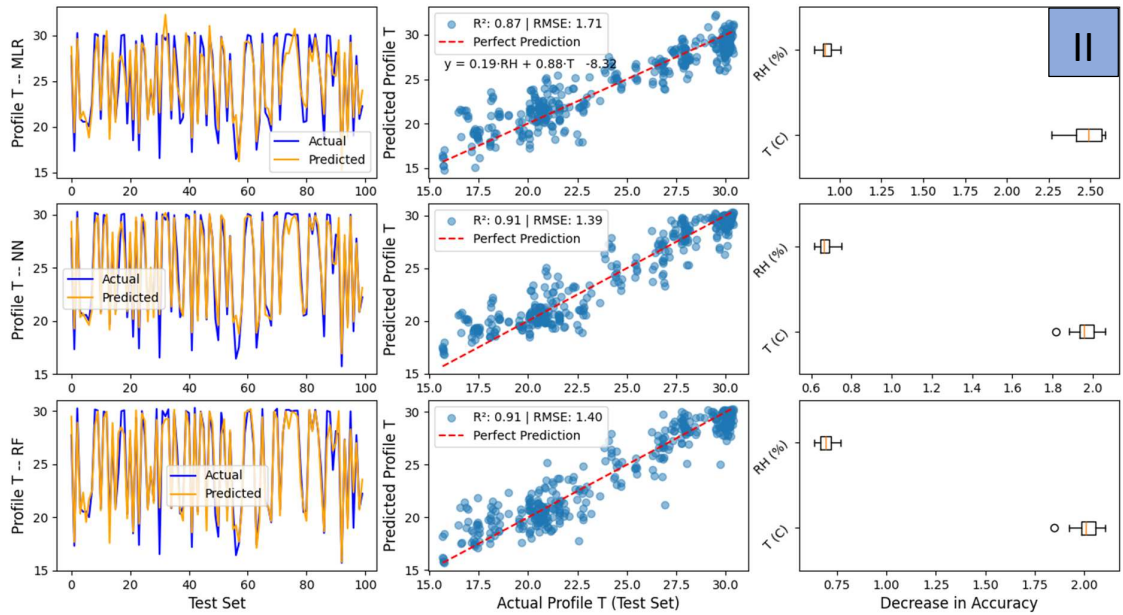

*Figure 2: This figure illustrates the predictive capabilities and feature importance for three models: (A) MLR, (B) NN and (C) RF. The left column presents time series plots comparing the actual (blue) and predicted (orange); Top-Soil Temperature (I) and 5-115 cm profile Temperature (II) values for each model, highlighting their performance over the test set. The central column features scatter plots of actual versus predicted values, along with the corresponding R-squared and RMSE metrics, offering a quantitative assessment of model accuracy. The right column showcases feature importance visualizations using permutation importance. In this method, the significance of a feature is evaluated by measuring the decrease in model performance (R-squared) after randomly shuffling its values, providing insights into the relative contribution of each feature to the model's predictive ability*

.

Table 1 - Model performance metrics for predicting effluent temperatures in recharge basins using MLR, NN, and RF models. The table reports the coefficient of determination ($R^2$) and root mean square error (RMSE, in °C) for both training and test sets

|  | Model | $R^2$ Train | RMSE-Train | $R^2$ Test | RMSE-Test |
|---|---|---|---|---|---|
| Top-Soil | MLR | 0.83 | 1.87 | 0.86 | 1.72 |
| Top-Soil | NN | 0.86 | 1.72 | 0.87 | 1.62 |
| Top-Soil | RF | 0.97 | 0.73 | 0.86 | 1.68 |
| Profile | MLR | 0.84 | 1.81 | 0.87 | 1.71 |
| Profile | NN | 0.89 | 1.49 | 0.91 | 1.39 |
| Profile | RF | 0.98 | 0.66 | 0.91 | 1.40 |

Table 2 Feature importance for ambient temperature (T, °C) and relative humidity (RH, %) in predicting effluent temperatures, assessed via permutation importance for MLR, NN, and RF models. Values represent the mean decrease in $R^2$ when each feature is randomly shuffled across the test set (10 repeats, random state 42), quantifying its contribution to model performance. Results are presented for both topsoil temperature at 5 cm depth and profile average temperature across 5–115 cm depths, based on the same 21-month dataset (June 3, 2022, to March 1, 2024) used in Table 1. Larger values indicate greater feature influence, with boxplots in Figure 2 I and II illustrating the variability of importance scores across permutations.

|  | Feature | MLR | NN | RF |
|---|---|---|---|---|
| Top-Soil | RH (%) | 0.93 | 0.76 | 0.74 |
| Top-Soil | T (C) | 2.58 | 2.23 | 2.20 |
| Profile | RH (%) | 0.92 | 0.67 | 0.69 |
| Profile | T (C) | 2.48 | 1.96 | 2.01 |

The results indicate that all three models MLR, NN, and RF exhibited strong predictive performance for both topsoil and profile average temperatures, with $R^2$ values ranging from 0.86 to 0.98 across training and test sets (Table 1). The RF model consistently achieved the highest accuracy, with test set $R^2$ values of 0.86 (topsoil) and 0.91 (profile) and RMSE values of 1.68°C and 1.40°C, respectively. However, the gap between its results and those of MLR is relatively modest, with MLR achieving test set $R^2$ values of 0.86 (topsoil) and 0.87 (profile) and RMSE values of 1.72°C and 1.71°C, respectively. This minimal difference in accuracy suggests that the added complexity of NN and RF models might not always yield substantial practical benefits in this specific application. Therefore, prioritizing a simpler model like MLR can be advantageous for real-world deployment where computational resources and ease of understanding are crucial. The MLR model offers simplicity and interpretability, making it easier to implement in operational systems. The denormalized MLR equations adopted are:

$$(Eq\ 2)\quad T_{\text{top-soil}} = 0.19 \cdot RH_{\text{ambient}} + 0.88 \cdot T_{\text{ambient}} - 8.05$$

and for profile average temperature,

$$(Eq\ 3)\quad T_{\text{profile}} = 0.19 \cdot RH_{\text{ambient}} + 0.88 \cdot T_{\text{ambient}} - 8.32.$$

These equations provide a practical tool for predicting effluent temperatures based on ambient temperature (T) and relative humidity (RH), balancing accuracy with operational feasibility.

### 3.2. Effluent Temperature Estimation

The MLR models were applied to a 10-year ambient dataset (2015-2025) to estimate effluent temperatures and assess their implications for recharge basin operations, with results visualized in Figure 3. Using equations 2 and 3, topsoil temperature at 5 cm depth and profile average temperature across 5-115 cm depths were predicted at 30-minute intervals.

Daily averages reveal pronounced seasonal cycles, with peaks in summer months (e.g., ~35 °C) and troughs in winter (~10 °C). Dynamic viscosity calculated from $T_{\text{profile}}$ using the Guzman-equation inversely correlated with temperature, with lower viscosities in warmer periods facilitating infiltration.

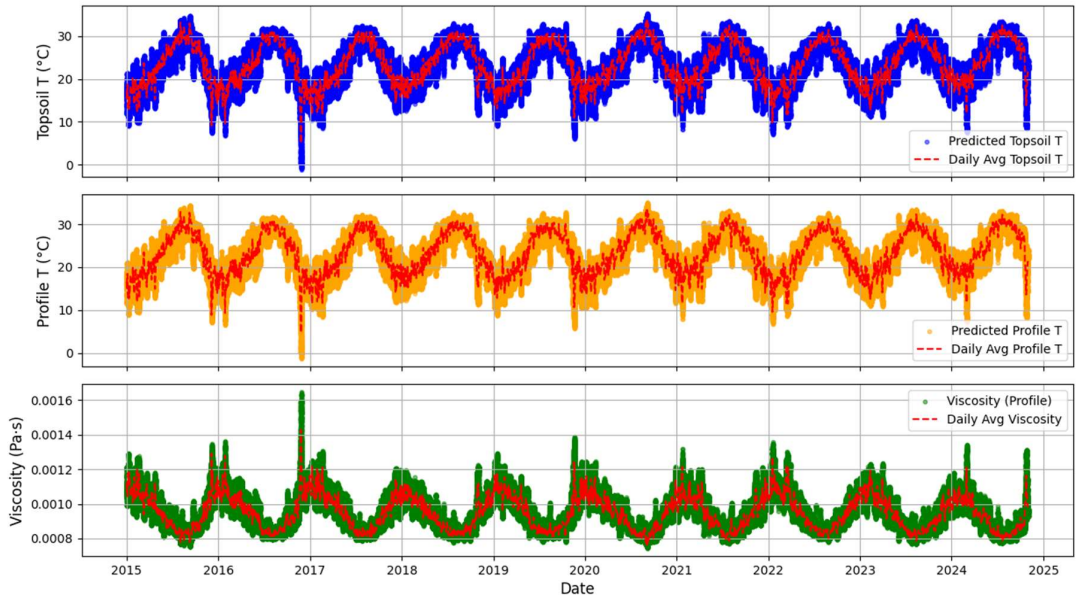

*Figure 3: Predicted effluent temperatures and dynamic viscosity over a 10-year period (2015–2025) at 30-minute intervals, derived from ambient temperature (T) and relative humidity (RH) data using MLR models. (A) Predicted topsoil temperature ($T_{top\text{-}soil}$) at 5 cm depth (blue scatter) with daily averages (red dashed line), (B) Predicted profile average temperature ($T_{profile}$) across 5–115 cm depths (orange scatter) with daily averages (red dashed line), (C) Dynamic viscosity of the profile derived from $T_{profile}$ using the Guzman-Andrade equation (green scatter) with daily averages (red dashed line). The scatter plots illustrate short-term variability, while daily averages highlight seasonal trends over the decade.*

The study demonstrates that ambient temperature and relative humidity are robust predictors of effluent temperature in recharge basins. The MLR model achieves accuracy comparable to more complex models, making it well-suited for operational applications where interpretability and computational efficiency are paramount. The strong seasonal temperature cycles observed have important implications for SAT operations. Warmer temperatures reduce water viscosity, increasing infiltration rates, while cooler temperatures have the opposite effect.

## 4. Conclusions

This research provides a practical and scalable framework for estimating effluent temperatures in recharge basins using ambient meteorological data. The MLR model offers a balance between accuracy and operational feasibility, supporting real-time monitoring and long-term planning of SAT systems. The derived equations and insights into seasonal temperature dynamics and viscosity variations are valuable for optimizing infiltration rates and improving the efficiency of wastewater reclamation processes.